\documentclass[11pt,a4paper]{article}
\usepackage[hyperref]{acl2017}
\usepackage{times}
\usepackage{latexsym}
\usepackage{url}
\usepackage{xspace}
\usepackage{multirow}
\usepackage{comment}
\usepackage{subcaption}
\usepackage[frozencache]{minted}
\usepackage{graphicx}
\usepackage{amsmath}

\newcommand{\Geo}{\textsc{Geo}\xspace}
\newcommand{\Jobs}{\textsc{Jobs}\xspace}
\newcommand{\Atis}{\textsc{Atis}\xspace}
\newcommand{\HS}{{\sc Hearthstone}\xspace}

\newcommand{\comp}{c}
\newcommand{\tok}{t}
\newcommand{\type}{\mathtt{T}}
\newcommand{\constructor}{\mathtt{C}}
\newcommand{\field}{\mathtt{F}}
\newcommand{\primValue}{\mathtt{y}}
\newcommand{\hForward}[2]{\overrightarrow{\mathbf{h}^{#1}_{#2}}}
\newcommand{\hBackward}[2]{\overleftarrow{\mathbf{h}^{#1}_{#2}}}
\newcommand{\vertState}{\mathbf{v}}
\newcommand{\horizState}{\mathbf{s}}
\newcommand{\hasChild}{z}
\newcommand{\vertLSTM}{\mathrm{LSTM}^{v}}
\newcommand{\horizLSTM}{\mathrm{LSTM}^{h}}
\newcommand{\embedding}{\mathbf{e}}
\newcommand{\ff}{f}
\newcommand{\projection}{\mathbf{W}}
\newcommand{\attVec}{\mathbf{c}}
\newcommand{\attScore}{q}
\newcommand{\attScoreVec}{\mathbf{\attScore}}
\newcommand{\attScoreRaw}{\attScore^{\mathrm{raw}}}
\newcommand{\attScoreComp}{\attScore^{\mathrm{comp}}}
\newcommand{\attWt}{a}
\newcommand{\attWtVec}{\mathbf{\attWt}}
\newcommand{\softmax}{\mathrm{softmax}}
\newcommand{\sigmoid}{\mathrm{sigmoid}}
\newcommand{\singular}{{\tt singular}\xspace}
\newcommand{\optional}{{\tt optional}\xspace}
\newcommand{\sequential}{{\tt sequential}\xspace}
\newcommand{\SeqToTree}{\textsc{Seq2Tree}\xspace}
\newcommand{\Nearest}{\textsc{Nearest}\xspace}
\newcommand{\LPN}{\textsc{LPN}\xspace}
\newcommand{\ASN}{\textsc{ASN}\xspace}
\newcommand{\SupAtt}{\textsc{SupAtt}\xspace}
\newcommand{\Prev}{\text{Previous Best}}

\aclfinalcopy % Uncomment this line for the final submission
\def\aclpaperid{707} %  Enter the acl Paper ID here

\title{Abstract Syntax Networks for Code Generation and Semantic Parsing}

\author{
  Maxim Rabinovich\thanks{\enskip Equal contribution.} \qquad Mitchell Stern\footnotemark[1] \qquad Dan Klein \\
  Computer Science Division \\
  University of California, Berkeley \\
  {\tt \{rabinovich,mitchell,klein\}@cs.berkeley.edu}
}

\date{}

\begin{document}
\maketitle

\begin{abstract}

Tasks like code generation and semantic parsing require mapping unstructured (or partially structured) inputs to well-formed, executable outputs. We introduce abstract syntax networks, a modeling framework for these problems. The outputs are represented as abstract syntax trees (ASTs) and constructed by a decoder with a dynamically-determined modular structure paralleling the structure of the output tree. On the benchmark \HS dataset for code generation, our model obtains 79.2 BLEU and 22.7\% exact match accuracy, compared to previous state-of-the-art values of 67.1 and 6.1\%. Furthermore,
we perform competitively on the \Atis, \Jobs, and \Geo semantic parsing datasets with no task-specific engineering. 
%we achieve state-of-the-art results on the \Atis and \Jobs semantic parsing datasets and competitive performance on the \Geo dataset.

%Code generation based on natural language specifications has long been a topic of interest in NLP. Recent sequence-to-sequence models for this task, while promising, do not account for programming language syntax and as a result miss out on crucial modeling and well-formedness constraints. In this work, we introduce a neural  translation architecture that allows for structure on both inputs and outputs and show  how to apply it to the problem of code generation. On the benchmark Hearthstone dataset, our model achieves 82 BLEU, compared to a previous 68 BLEU state-of-the-art. Furthermore, we show that an off-the-shelf specialization of our approach achieves state-of-the-art semantic parsing performance on the \Jobs and \Atis datasets, and is competitive on the \Geo dataset.
\end{abstract}

\section{Introduction}

\begin{figure}[t]
\begin{minipage}[contentpos=t]{0.5\linewidth}
\includegraphics[trim={0cm 0cm 0cm 0.6cm},clip,width=\linewidth]{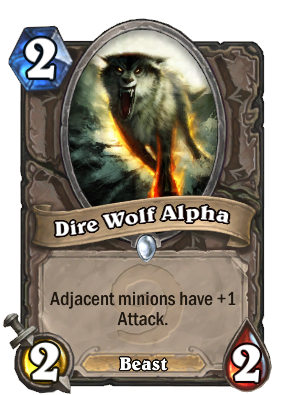}
\end{minipage}%
\begin{minipage}[contentpos=t]{0.5\linewidth}
\begin{minted}[fontsize=\scriptsize]{text}
name: [
  'D', 'i', 'r', 'e', ' ', 
  'W', 'o', 'l', 'f', ' ', 
  'A', 'l', 'p', 'h', 'a']
cost: ['2']
type: ['Minion']
rarity: ['Common']
race: ['Beast']
class: ['Neutral']
description: [
  'Adjacent', 'minions', 'have',
  '+', '1', 'Attack', '.']
health: ['2']
attack: ['2']
durability: ['-1']
\end{minted}
\end{minipage}
\begin{minted}[fontsize=\scriptsize]{python}
class DireWolfAlpha(MinionCard):
  def __init__(self):
    super().__init__(
      "Dire Wolf Alpha", 2, CHARACTER_CLASS.ALL, 
      CARD_RARITY.COMMON, minion_type=MINION_TYPE.BEAST)
  def create_minion(self, player):
    return Minion(2, 2, auras=[
      Aura(ChangeAttack(1), MinionSelector(Adjacent()))
    ])
\end{minted}
\caption{Example code for the ``Dire Wolf Alpha'' Hearthstone card.\label{fig:card-example}}
\end{figure}

\begin{figure}[!t]
\footnotesize
\begin{verbatim}
show me the fare from ci0 to ci1 

lambda $0 e 
  ( exists $1 ( and ( from $1 ci0 ) 
                    ( to $1 ci1 ) 
                    ( = ( fare $1 ) $0 ) ) )
\end{verbatim}
\caption{Example of a query and its logical form from the \Atis dataset. The {\tt ci0} and {\tt ci1} tokens are entity abstractions introduced in preprocessing~\citep{Dong16Semantic}.\label{fig:lambda-example}}
\end{figure}

% New version (02/04/17).
%Recently, there has been great interest in tasks like semantic parsing and code generation, where a structured output with significant well-formedness constraints must be predicted based on an unstructured (or partially structured) input with a textual component. Unlike in tasks such as constituency parsing, dependency parsing, or (some models of) translation, the output structure in semantic parsing and code generation does not directly parallel the structure of the input---making it more difficult to leverage.

\noindent Tasks like semantic parsing and code generation are challenging in part because they are \emph{structured} (the output must be well-formed) but not \emph{synchronous} (the output structure diverges from the input structure).

Sequence-to-sequence models have proven effective for both tasks~\citep{Dong16Semantic,Ling16Code}, using encoder-decoder frameworks to exploit the sequential structure on both the input and output side. Yet these approaches do not account for much richer structural constraints on outputs---including well-formedness, well-typedness, and executability. The well-formedness case is of particular interest, since it can readily be enforced by representing outputs as abstract syntax trees (ASTs)~\citep{Aho06Compilers}, an approach that can be seen as a much lighter weight version of CCG-based semantic parsing~\citep{Zettlemoyer05Semantic}.

In this work, we introduce {\it abstract syntax networks} (ASNs), an extension of the standard encoder-decoder framework utilizing a modular decoder whose submodels are composed to natively generate ASTs in a top-down manner. The decoding process for any given input follows a dynamically chosen mutual recursion between the modules, where the structure of the tree being produced mirrors the call graph of the recursion. We implement this process using a decoder model built of many submodels, each associated with a specific construct in the AST grammar and invoked when that construct is needed in the output tree. As is common with neural approaches to structured prediction~\citep{Chen14Fast,Vinyals15Foreign}, our decoder proceeds greedily and accesses not only a fixed encoding but also an attention-based representation of the input~\citep{Bahdanau14Attention}.

Our model significantly outperforms previous architectures for code generation and obtains competitive or state-of-the-art results on a suite of semantic parsing benchmarks. On the \HS dataset for code generation, we achieve a token BLEU score of 79.2 and an exact match accuracy of 22.7\%, greatly improving over the previous best results of 67.1 BLEU and 6.1\% exact match~\citep{Ling16Code}.

The flexibility of ASNs makes them readily applicable to other tasks with minimal adaptation. We illustrate this point with a suite of semantic parsing experiments. On the \Jobs dataset, we improve on previous state-of-the-art, achieving 92.9\% exact match accuracy as compared to the previous record of 90.7\%. Likewise, we perform competitively on the \Atis and \Geo datasets, matching or exceeding the exact match reported by~\citet{Dong16Semantic}, though not quite reaching the records held by the best previous semantic parsing approaches~\citep{Wang14Morpho}.

\begin{figure*}[t]
\setbox1=\hbox{\includegraphics[trim={1.5cm 4.5cm 2.75cm 1.5cm}, clip, width=0.5\linewidth]{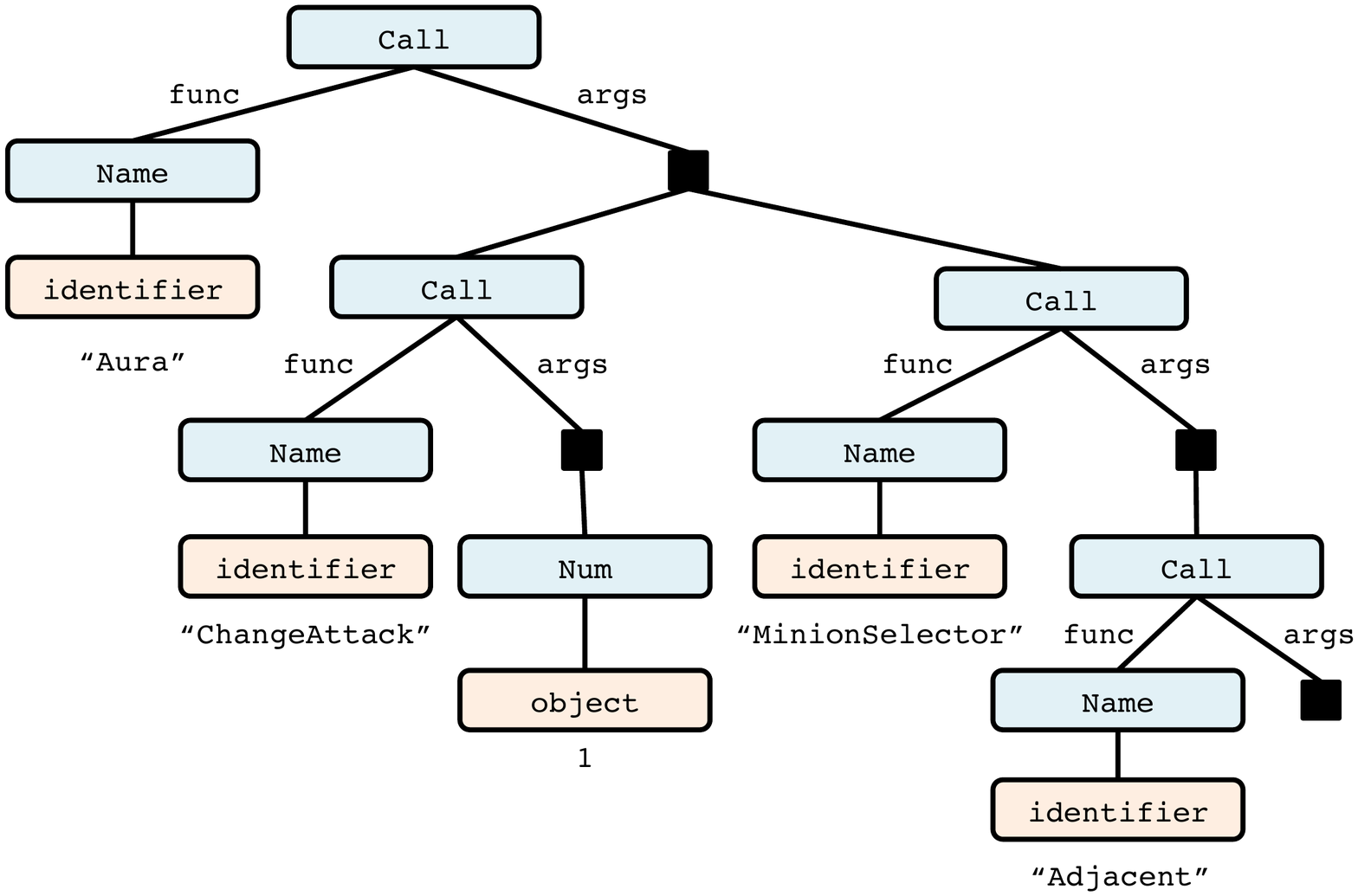}}
\subcaptionbox{\label{fig:ast-whole-example}The root portion of the AST.}{\raisebox{\dimexpr\ht1-\height}{\includegraphics[trim={1.75cm 10cm 4cm 1.5cm}, clip, width=0.5\linewidth]{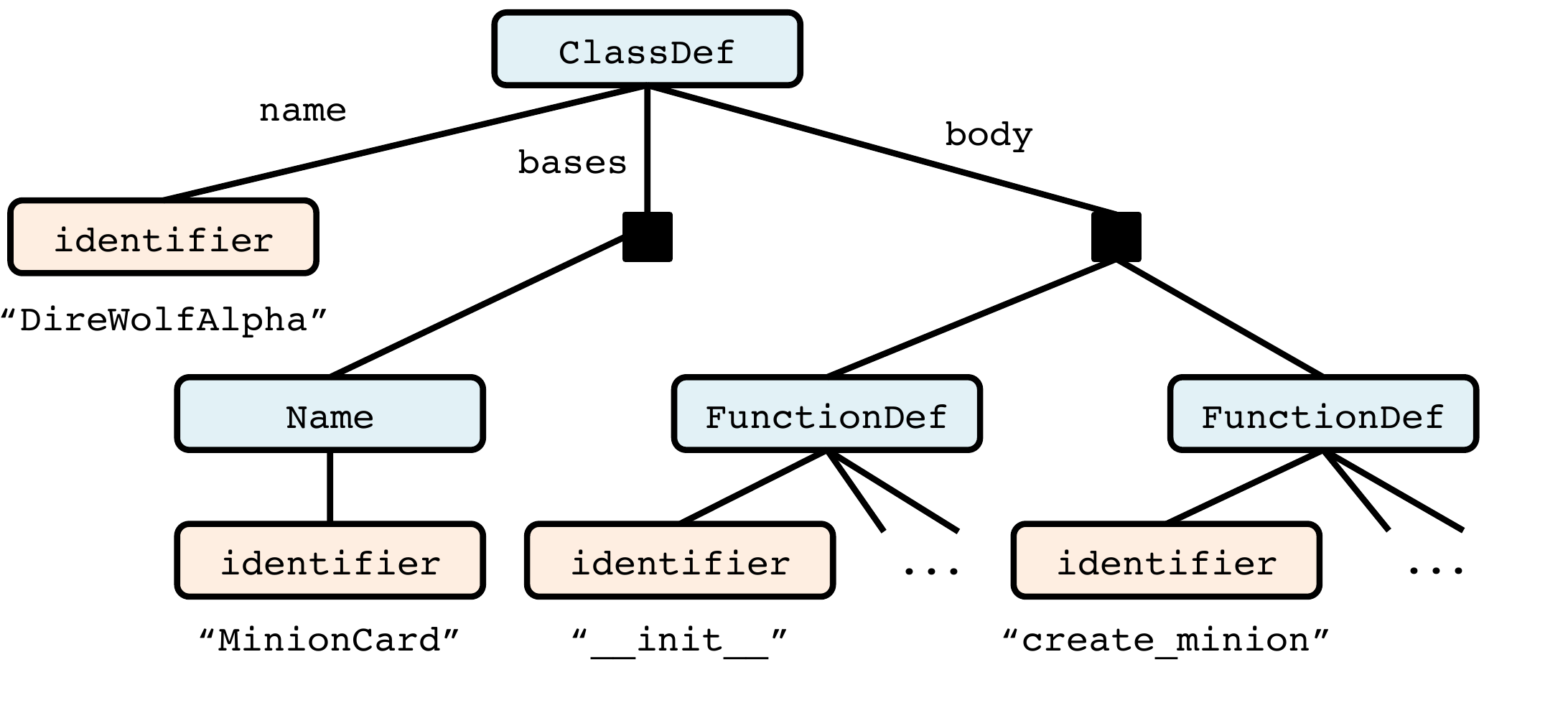}}}
\hfill
\subcaptionbox{\label{fig:ast-part-example} Excerpt from the same AST, corresponding to the code snippet \mintinline[fontsize=\scriptsize]{python}|Aura(ChangeAttack(1),MinionSelector(Adjacent()))|.}{\includegraphics[trim={1.5cm 4.5cm 2.75cm 1.5cm}, clip, width=0.5\linewidth]{ast_part.pdf}}
\caption{Fragments from the abstract syntax tree corresponding to the example code in Figure~\ref{fig:card-example}. Blue boxes represent composite nodes, which expand via a constructor with a prescribed set of named children. Orange boxes represent primitive nodes, with their corresponding values written underneath. Solid black squares correspond to constructor fields with \sequential cardinality, such as the body of a class definition (Figure~\ref{fig:ast-whole-example}) or the arguments of a function call (Figure~\ref{fig:ast-part-example}).}
\end{figure*}

\subsection{Related work}

% Dong and Lapata semantic parsing paper [https://arxiv.org/pdf/1601.01280v2.pdf]
% ICLR 2017 submissions on tree decoders

% Direct antecedents 1: encoder-decoder architectures, sequence-to-sequence models.
Encoder-decoder architectures, with and without attention, have been applied successfully both to sequence prediction tasks like machine translation and to tree prediction tasks like constituency parsing~\citep{Cross16Constituency,Dyer16RNNG,Vinyals15Foreign}. In the latter case, work has focused on making the task look like sequence-to-sequence prediction, either by flattening the output tree~\citep{Vinyals15Foreign} or by representing it as a sequence of construction decisions~\citep{Cross16Constituency,Dyer16RNNG}. Our work differs from both in its use of a recursive top-down generation procedure.

% Direct antecedents 2: neural approaches to semantic parsing, code generation. Include ICRL 2017 submissions (synthetic data, simpler IFTTT trees).
%
% Less direct antecedents: modeling code (various Tarlow), program induction (Liang et al '10, Gulwani et al '13). 
\citet{Dong16Semantic} introduced a sequence-to-sequence approach to semantic parsing, including a limited form of top-down recursion, but without the modularity or tight coupling between output grammar and model characteristic of our approach. 
%Previously, most work on semantic parsing had used structured prediction approaches~\citep{Kwiatowski13OnTheFly,Zettlemoyer05Semantic,Zhao15Incremental}; that line of work remains state-of-the-art for the \Geo dataset.
% MS: approach, approaches

Neural (and probabilistic) modeling of code, including for prediction problems, has a longer history. \citet{Allamanis15Bimodal} and~\citet{ Maddison14SourceCode} proposed modeling code with a neural language model, generating concrete syntax trees in left-first depth-first order, focusing on metrics like perplexity and applications like code snippet retrieval. More recently, \citet{Shin17VAE} attacked the same problem using a grammar-based variational autoencoder with top-down generation similar to ours instead. Meanwhile, a separate line of work has focused on the problem of program induction from input-output pairs~\citep{Balog16DeepCoder,Liang10LearningPrograms,Menon13MLProgramming}.
%Some work on modeling or retrieving, rather than predicting, code has used real source code~\citep{Allamanis15Bimodal,Maddison14SourceCode}. Most prior work, however, has focused on a combination of synthetic data and simple code snippets like those from the \IFTTT dataset~\citep{Quirk15IFTTT}. 

The prediction framework most similar in spirit to ours is the doubly-recurrent decoder network introduced by~\citet{AlvarezMelis17Doubly}, which propagates information down the tree using a vertical LSTM and between siblings using a horizontal LSTM. Our model differs from theirs in using a separate module for each grammar construct and learning separate vertical updates for siblings when the AST labels require all siblings to be jointly present; we do, however, use a horizontal LSTM for nodes with {\it variable} numbers of children. The differences between our models reflect not only design decisions, but also differences in data---since ASTs have labeled nodes and labeled edges, they come with additional structure that our model exploits. 
%Our work also differs from theirs in using the \HS dataset, whose code is significantly more complex than the \IFTTT dataset~\citet{AlvarezMelis17Doubly} consider.

Apart from ours, the best results on the code-generation task associated with the \HS dataset are based on a sequence-to-sequence approach to the problem~\citep{Ling16Code}. Abstract syntax networks greatly improve on those results. 

Previously,~\citet{Andreas16Neural} introduced neural module networks (NMNs) for visual question answering, with modules corresponding to linguistic substructures within the input query. The primary purpose of the modules in NMNs is to compute deep features of images in the style of convolutional neural networks (CNN). These features are then fed into a final decision layer. In contrast to the modules we describe here, NMN modules do not make decisions about what to generate or which modules to call next, nor do they maintain recurrent state. 

% Whereas their architecture composes models in a static order determined by the dependency parse of the input, the modular structure instantiated by abstract syntax networks is determined dynamically during the decoding process.

% Whereas the compositional structure of NMNs is determined in a static manner according to the dependency parse of the input, the modular structure instantiated by abstract syntax networks is determined dynamically during the decoding process.

% More tangential: tree LSTMs.
%Some recent work has sought to generalize LSTMs to work natively with trees, resulting in tree-shaped LSTMs~\citep{Tai15TreeLSTM}, but these approaches use a fixed tree-structure and pass messages from the leaves up. Our abstract syntax networks generate the tree structure dynamically and propagate state from the top down.

\section{Data Representation}

\subsection{Abstract Syntax Trees}

% High-level recap of AST structures: 3 general types of nodes, namely composite nodes for language constructs, sequence nodes for bodies, and leaves for values.

Our model makes use of the Abstract Syntax Description Language (ASDL) framework~\citep{Wang97ASDL}, which represents code fragments as trees with typed nodes. {\it Primitive types} correspond to atomic values, like integers or identifiers. Accordingly, {\it primitive nodes} are annotated with a primitive type and a value of that type---for instance, in Figure~\ref{fig:ast-whole-example}, the \texttt{identifier} node storing \texttt{"create\_minion"} represents a function of the same name.

{\it Composite types} correspond to language constructs, like expressions or statements. Each type has a collection of {\it constructors}, each of which specifies the particular language construct a node of that type represents. Figure~\ref{fig:python-grammar-fragment} shows constructors for the statement ({\tt stmt}) and expression ({\tt expr}) types. The associated language constructs include function and class definitions, return statements, binary operations, and function calls.

\begin{figure}[h]
\begin{minted}[fontsize=\footnotesize]{text}
primitive types: identifier, object, ...

stmt
 = FunctionDef(
    identifier name, arg* args, stmt* body)
 | ClassDef(
    identifier name, expr* bases, stmt* body)
 | Return(expr? value)
 | ...

expr
 = BinOp(expr left, operator op, expr right)
 | Call(expr func, expr* args)
 | Str(string s)
 | Name(identifier id, expr_context ctx)
 | ...

...
\end{minted}
\caption{A simplified fragment of the Python ASDL grammar.\footnotemark}
%It contains top-level composite types such as \texttt{stmt} and \texttt{expr}, as well as primitive types such as \texttt{identifier} and \texttt{object}. Composite types have constructors like \texttt{FunctionDef} or \texttt{BinOp} with named fields designating their children. Each field has a type and a cardinality---unannotated types correspond to \singular fields, whereas \texttt{type*} denotes a \sequential field and \texttt{type?} denotes an \optional field.}
\label{fig:python-grammar-fragment}
\end{figure}

\footnotetext{The full grammar can be found online on the documentation page for the Python {\tt ast} module: \url{https://docs.python.org/3/library/ast.html\#abstract-grammar}}

Composite types enter syntax trees via {\it composite nodes}, annotated with a composite type and a choice of constructor specifying how the node expands. The root node in Figure~\ref{fig:ast-whole-example}, for example, is a composite node of type \texttt{stmt} that represents a class definition and therefore uses the {\tt ClassDef} constructor. In Figure~\ref{fig:ast-part-example}, on the other hand, the root uses the {\tt Call} constructor because it represents a function call.

Children are specified by named and typed {\it fields} of the constructor, which have {\it cardinalities} of \singular, \optional, or \sequential. By default, fields have \singular cardinality, meaning they correspond to exactly one child. For instance, the {\tt ClassDef} constructor has a \singular name field of type {\tt identifier}. Fields of \optional cardinality are associated with zero or one children, while fields of \sequential cardinality are associated with zero or more children---these are designated using {\tt ?} and {\tt *} suffixes in the grammar, respectively. Fields of \sequential cardinality are often used to represent statement blocks, as in the {\tt body} field of the {\tt ClassDef} and {\tt FunctionDef} constructors.

\begin{figure*}[t]
\centering
\begin{subfigure}{0.45\linewidth}
\centering
\includegraphics[scale=0.6]{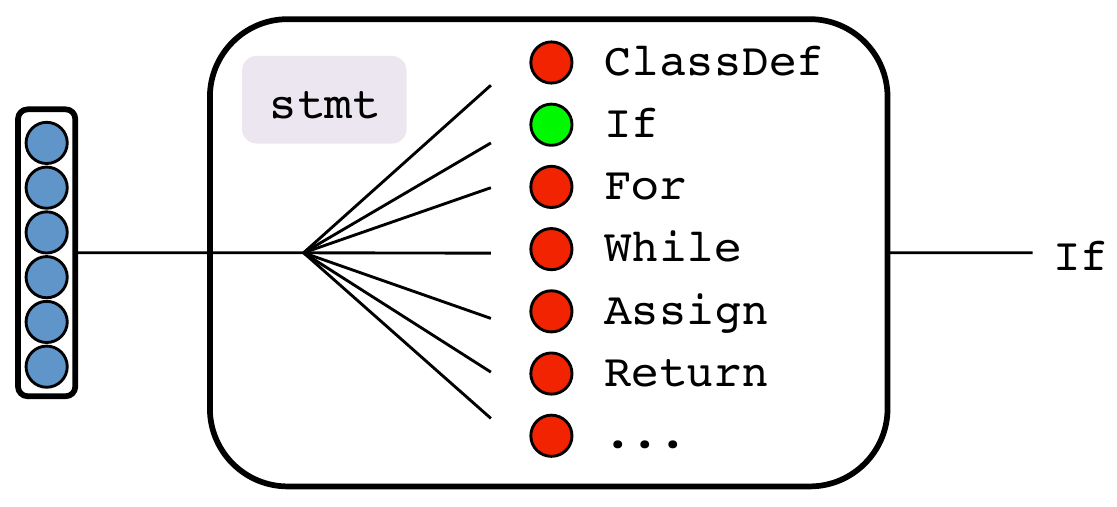}
\caption{A composite type module choosing a constructor for the corresponding type.\label{subfig:composite}}
\end{subfigure}%
\hspace{0.3in}\begin{subfigure}{0.45\linewidth}
\centering
\includegraphics[scale=0.6]{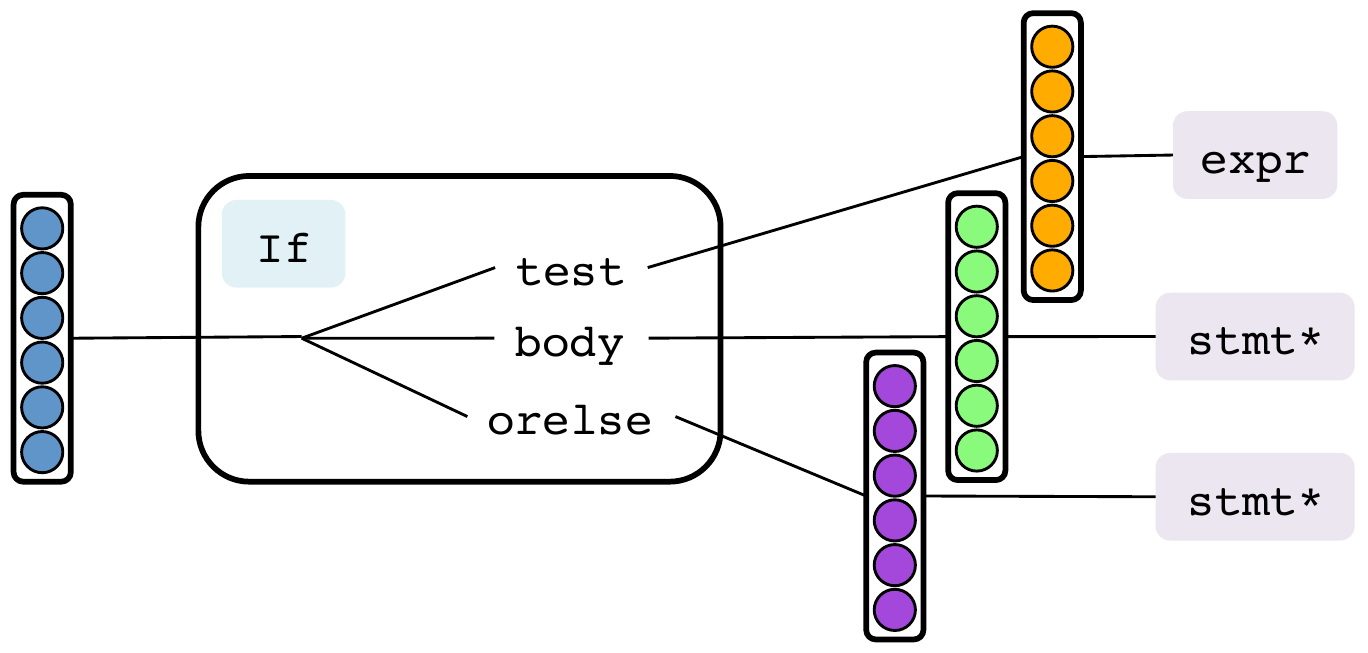}
\caption{A constructor module computing updated vertical LSTM states.\label{subfig:constructor}}
\end{subfigure}
\begin{subfigure}{0.45\linewidth}
\centering
\includegraphics[scale=0.6]{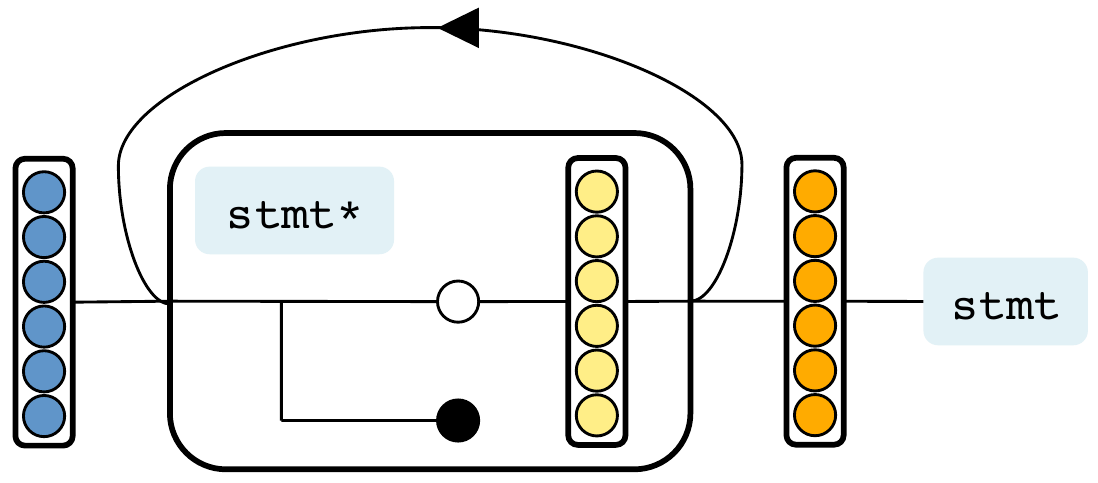}
\caption{A constructor field module (\sequential cardinality) generating children to populate the field. At each step, the module decides whether to generate a child and continue (white circle) or stop (black circle). \label{subfig:sequence}}
\end{subfigure}%
\hspace{0.3in}\begin{subfigure}{0.45\linewidth}
\centering
\includegraphics[scale=0.6]{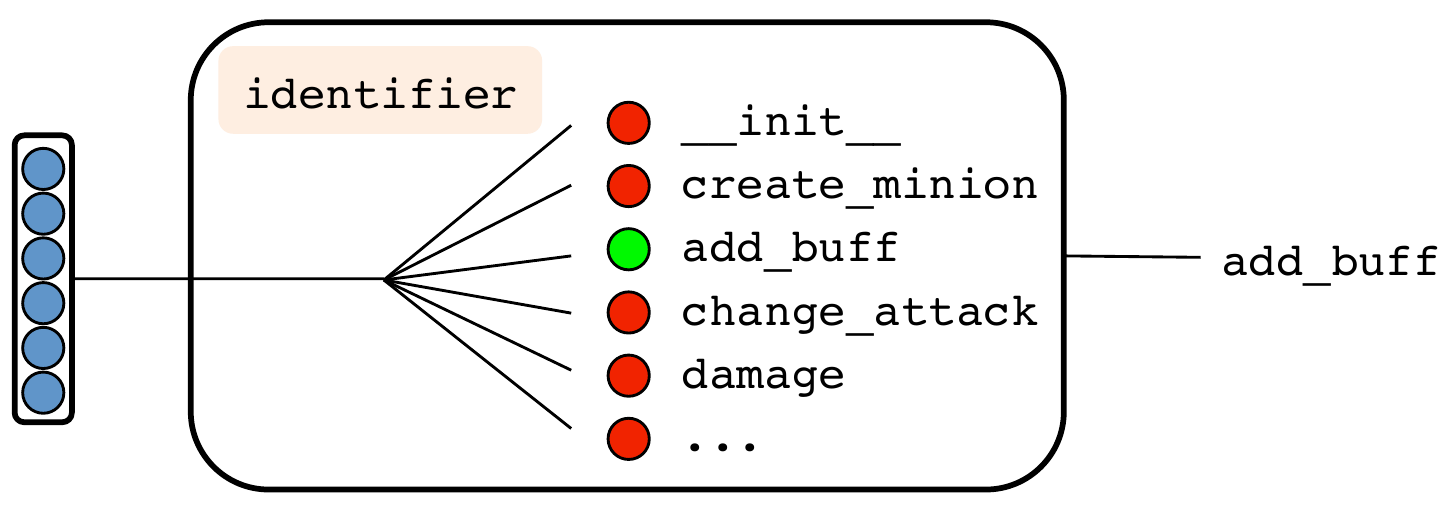}
\caption{A primitive type module choosing a value from a closed list.\label{subfig:primitive}}
\end{subfigure}
\caption{The module classes constituting our decoder. For brevity, we omit the cardinality modules for \singular and \optional cardinalities.\label{fig:modules}}
\end{figure*}

%The root node in Figure~\ref{fig:ast-part-example}, labeled {\tt Call}, is an example of a composite node of type \texttt{expr} that represents a function call and therefore uses the {\tt Call} constructor. This constructor has a {\tt func} field of \singular cardinality for the name of the function being called, as well as an {\tt args} field of \sequential cardinality corresponding to a variable number of arguments.

%For example, a composite node with type \texttt{stmt} that represents a function definition would use the {\tt FunctionDef} constructor, which has a name field of type {\tt identifier} and a body field of type {\tt stmt}. Since a function only has a single name but consists of a variable number of statements, the cardinality of the name field is \singular, whereas the cardinality of the body field is \sequential.

%Figure~\ref{fig:py-grammar} shows the ASDL specification for the grammar of Python we used.

% Application to semantic parsing.

The grammars needed for semantic parsing can easily be given ASDL specifications as well, using primitive types to represent variables, predicates, and atoms and composite types for standard logical building blocks like lambdas and counting (among others). Figure~\ref{fig:lambda-example} shows what the resulting $\lambda$-calculus trees look like. The ASDL grammars for both $\lambda$-calculus and Prolog-style logical forms are quite compact, as Figures~\ref{fig:prolog-grammar} and~\ref{fig:lambda-grammar} in the appendix show.

\subsection{Input Representation}

We represent inputs as collections of named components, each of which consists of a sequence of tokens. In the case of semantic parsing, inputs have a single component containing the query sentence. In the case of \HS, the card's name and description are
represented as sequences of characters and tokens, respectively, while categorical attributes are represented as single-token sequences. For \HS, we restrict our input and output vocabularies to values that occur more than once in the training set.

% TODO: Integrate semantic parsing more closely into the discussion.

% State updates, attention, etc.
\section{Model Architecture}

%Our model uses an encoder-decoder architecture with hierarchical attention. The encoder uses bidirectional LSTMs to embed each component and a feedforward network to combine them. The decoder generates trees recursively in a top-down manner, using a vertical LSTM to propagate information from parents to children. Component- and token-level attention is applied over the input at each step of the decoding process.

Our model uses an encoder-decoder architecture with hierarchical attention. The key idea behind our approach is to structure the decoder as a collection of mutually recursive modules. The modules correspond to elements of the AST grammar and are composed together in a manner that mirrors the structure of the tree being generated. A vertical LSTM state is passed from module to module to propagate information during the decoding process.

The encoder uses bidirectional LSTMs to embed each component and a feedforward network to combine them. Component- and token-level attention is applied over the input at each step of the decoding process.

We train our model using negative log likelihood as the loss function. The likelihood encompasses terms for all generation decisions made by the decoder.

% TODO: Overview of the modularity and its relationship to the shape of the output.

\subsection{Encoder}\label{subsec:encoder}

Each component $\comp$ of the input is encoded using a component-specific bidirectional LSTM. This results in forward and backward token encodings ($\hForward{c}{}, \hBackward{c}{}$) that are later used by the attention mechanism. To obtain an encoding of the input as a whole for decoder initialization, we concatenate the final forward and backward encodings of each component into a single vector and apply a linear projection.

\subsection{Decoder Modules}\label{subsec:decoder-modules}

The decoder decomposes into several classes of modules, one per construct in the grammar, which we discuss in turn. Throughout, we let $\vertState$ denote the current vertical LSTM state, and use $\ff$ to represent a generic feedforward neural network. LSTM updates with hidden state $\mathbf{h}$ and input $\mathbf{x}$ are notated as $\mathrm{LSTM}(\mathbf{h}, \mathbf{x})$.

\paragraph{Composite type modules} Each composite type $\type$ has a corresponding module whose role is to select among the constructors $\constructor$ for that type. As Figure~\ref{subfig:composite} exhibits, a composite type module receives a vertical LSTM state $\vertState$ as input and applies a feedforward network $\ff_{\type}$ and a softmax output layer to choose a constructor:
\begin{equation*}
  p\left(\constructor~|~\type,\vertState\right) = \big[\softmax\left(\ff_{\type}\left(\vertState\right)\right)\!\big]_{\constructor} .
\end{equation*}
Control is then passed to the module associated with constructor $\constructor$.

\paragraph{Constructor modules} Each constructor $\constructor$ has a corresponding module whose role is to compute an intermediate vertical LSTM state $\vertState_{u,\field}$ for each of its fields $\field$ whenever $\constructor$ is chosen at a composite node $u$. 

%To implement this update, an embedding for a given constructor field $\field$ is concatenated with an attention vector, then run through a feedforward network to obtain a context-dependent field embedding. The resulting vector is fed as input to the vertical LSTM to obtain the new state for the corresponding child.

For each field $\field$ of the constructor, an embedding $\embedding_{\field}$ is concatenated with an attention-based context vector $\attVec$ and fed through a feedforward neural network $\ff_{\constructor}$ to obtain a context-dependent field embedding:
\begin{equation*}
 \tilde{\embedding}_{\field} = \ff_{\constructor}\left(\embedding_{\field},~\attVec\right) .
\end{equation*}
An intermediate vertical state for the field $\field$ at composite node $u$ is then computed as
\begin{equation*}
    \vertState_{u,\field} = \vertLSTM\left(\vertState_{u},~\tilde{\embedding}_{\field}\right).
\end{equation*}
Figure~\ref{subfig:constructor} illustrates the process, starting with a single vertical LSTM state and ending with one updated state per field.

\paragraph{Constructor field modules} Each field $\field$ of a constructor has a corresponding module whose role is to determine the number of children associated with that field and to propagate an updated vertical LSTM state to them. In the case of fields with \singular cardinality, the decision and update are both vacuous, as exactly one child is always generated. Hence these modules forward the field vertical LSTM state $\vertState_{u,\field}$ unchanged to the child $w$ corresponding to $\field$:
\begin{equation}\label{eq:vert-singular}
  \vertState_{w} = \vertState_{u,\field} .
\end{equation}
Fields with \optional cardinality can have either zero or one children; this choice is made using a feedforward network applied to the vertical LSTM state:
\begin{equation}\label{eq:haschild-optional}
  p(\hasChild_{\field} = 1~|~\vertState_{u,\field}) = \sigmoid\left(\ff_{\field}^{\mathrm{gen}}(\vertState_{u,\field})\right) .
\end{equation}
If a child is to be generated, then as in~\eqref{eq:vert-singular}, the state is propagated forward without modification.

In the case of \sequential fields, a horizontal LSTM is employed for both child decisions and state updates. We refer to Figure~\ref{subfig:sequence} for an illustration of the recurrent process. After being initialized with a transformation of the vertical state, $\horizState_{\field,0} = \projection_{\field} \vertState_{u,\field}$, the horizontal LSTM iteratively decides whether to generate another child by applying a modified form of~\eqref{eq:haschild-optional}:
\begin{align*}
  & p\left(\hasChild_{\field,i} = 1~|~\horizState_{\field,i-1},~\vertState_{u,\field}\right) = \nonumber \\
  & \qquad\sigmoid\left(\ff_{\field}^{\mathrm{gen}}\left(\horizState_{\field,i-1},~\vertState_{u,\field}\right)\right) .
\end{align*}
If $\hasChild_{\field,i} = 0$, generation stops and the process terminates, as represented by the solid black circle in Figure~\ref{subfig:sequence}. Otherwise, the process continues as represented by the white circle in Figure~\ref{subfig:sequence}. In that case, the horizontal state $\horizState_{u,i-1}$ is combined with the vertical state $\vertState_{u,\field}$ and an attention-based context vector $\attVec_{\field,i}$ using a feedforward network $\ff_{\field}^{\mathrm{update}}$ to obtain a joint context-dependent encoding of the field $\field$ and the position $i$:
\begin{equation*}
    \tilde{\embedding}_{\field,i} = \ff_{\field}^{\mathrm{update}}(\vertState_{u,\field},~\horizState_{u,i-1},~ \attVec_{\field,i}) .
\end{equation*}
The result is used to perform a vertical LSTM update for the corresponding child $w_{i}$:
\begin{equation*}
  \vertState_{w_i} = \vertLSTM(\vertState_{u,\field},~\tilde{\embedding}_{\field,i}) .
\end{equation*}
Finally, the horizontal LSTM state is updated using the same field-position encoding, and the process continues:
\begin{equation*}
  \horizState_{u,i} = \horizLSTM(\horizState_{u,i-1},~\tilde{\embedding}_{\field,i}) .
\end{equation*}

% Paragraph on horizontal LSTM details.

\paragraph{Primitive type modules} Each primitive type $\type$ has a corresponding module whose role is to select among the values $\primValue$ within the domain of that type. Figure~\ref{subfig:primitive} presents an example of the simplest form of this selection process, where the value $\primValue$ is obtained from a closed list via a softmax layer applied to an incoming vertical LSTM state:
\begin{equation*}
  p\left(\primValue~|~\type,\vertState\right) = \big[\softmax\left(\ff_{\type}\left(\vertState\right)\right)\!\big]_{\text{\footnotesize $\primValue$}} .
\end{equation*}

% Short paragraph on string module.
Some string-valued types are open class, however. To deal with these, we allow generation both from a closed list of previously seen values, as in Figure~\ref{subfig:primitive}, and synthesis of new values. Synthesis is delegated to a character-level LSTM language model~\citep{Bengio03Neural}, and part of the role of the primitive module for open class types is to choose whether to synthesize a new value or not. During training, we allow the model to use the character LSTM only for unknown strings but include the log probability of that binary decision in the loss in order to ensure the model learns when to generate from the character LSTM. 

\subsection{Decoding Process}\label{subsec:decoding-process}

% TODO: Need to discuss root type decision somewhere, either in this section or as a short "root module" in the previous section.

The decoding process proceeds through mutual recursion between the constituting modules, where the syntactic structure of the output tree mirrors the call graph of the generation procedure. At each step, the active decoder module either makes a generation decision, propagates state down the tree, or both. 

To construct a composite node of a given type, the decoder calls the appropriate composite type module to obtain a constructor and its associated module. That module is then invoked to obtain updated vertical LSTM states for each of the constructor's fields, and the corresponding constructor field modules are invoked to advance the process to those children.

This process continues downward, stopping at each primitive node, where a value is
generated but no further recursion is carried out.

\subsection{Attention}\label{subsec:attention}

Following standard practice for sequence-to-sequence models, we compute a raw bilinear attention score $\attScoreRaw_{\tok}$ for each token $\tok$ in the input using the decoder's current state $\mathbf{x}$ and the token's encoding $\embedding_{\tok}$:
\begin{equation*}
  \attScoreRaw_{\tok} = \embedding_{\tok}^{\top} \projection \mathbf{x} .
\end{equation*}
The current state $\mathbf{x}$ can be either the vertical LSTM state in isolation or a concatentation of the vertical LSTM state and either a horizontal LSTM state or a character LSTM state (for string generation). Each submodule that computes attention does so using a separate matrix $\projection$.

A separate attention score $\attScoreComp_{\comp}$ is computed for each component of the input, independent of its content:
\begin{equation*}
  \attScoreComp_{\comp} = \mathbf{w}_{\comp}^{\top} \mathbf{x} .
\end{equation*}

The final token-level attention scores are the sums of the raw token-level scores and the corresponding component-level scores:
\begin{equation*}
\attScore_{\tok} = \attScoreRaw_{\tok} + \attScoreComp_{\comp(\tok)} ,    
\end{equation*}
where $\comp(\tok)$ denotes the component in which token $\tok$ occurs. The attention weight vector $\attWtVec$ is then computed using a softmax:
\begin{equation*}
  \attWtVec = \softmax\left(\attScoreVec\right) .
\end{equation*}
Given the weights, the attention-based context is given by:
\begin{equation*}
  \mathbf{c} = \sum_{\tok} \attWt_{\tok} \embedding_{\tok} .
\end{equation*}

Certain decision points that require attention have been highlighted in the description above; however, in our final implementation we made attention available to the decoder at all decision points.

\paragraph{Supervised Attention}

In the datasets we consider, partial or total copying of input tokens into primitive nodes is quite common. Rather than providing an explicit copying mechanism~\citep{Ling16Code}, we instead generate alignments where possible to
define a set of tokens on which the attention at a given primitive node should be concentrated.\footnote{Alignments are generated using an exact string match heuristic that also included some limited normalization, primarily splitting of special characters, undoing camel case, and lemmatization for the semantic parsing datasets.} If no matches are found, the corresponding set of tokens is taken to be the whole input.

The attention supervision enters the loss through a term that encourages the final attention weights to be concentrated on the specified subset. Formally, if the matched subset of component-token pairs is $S$, the loss term associated with the supervision would be
\begin{equation}\label{eq:attention-loss}
    \log{\sum_{\tok} \exp\left(a_{\tok}\right)} - 
    \log{\sum_{\tok \in S} \exp\left(a_{\tok}\right)},
\end{equation}
where $a_{\tok}$ is the attention weight associated with token $\tok$, and the sum in the first term ranges over all tokens in the input. The loss in~\eqref{eq:attention-loss} can be interpreted as the negative log probability of attending to some token in $S$.

\section{Experimental evaluation}

\begin{comment}
\begin{table}[t]
\centering
\begin{tabular}{clc}
\toprule
& Method & Accuracy (\%) \\ \midrule
\multirow{4}{*}{\Atis} & $\Prev^{\dagger}$ & 84.6 \\
 & \SeqToTree & 84.6 \\
 & \ASN & {\bf 85.3} \\
 &~~ + \SupAtt & {\bf 85.9} \\ \midrule
\multirow{4}{*}{\Geo} & $\Prev^{\ddagger}$ & {\bf 89.0} \\
 & \SeqToTree & 87.1 \\
 & \ASN & 85.7 \\
 &~~ + \SupAtt & 87.1 \\ \midrule
\multirow{4}{*}{\Jobs} & $\Prev^{\dagger\dagger}$ & 90.7 \\
 & \SeqToTree & 90.0 \\
 & \ASN & {\bf 91.4} \\
 &~~ + \SupAtt & {\bf 92.9}
\end{tabular}
\caption{Accuracies for the semantic parsing task. \SupAtt refers to the system with supervised attention mentioned in Section~\ref{subsec:attention}. \SeqToTree refers to the system of~\citet{Dong16Semantic}. \newline
$\dagger$~\citep{Zettlemoyer07Online},~ $\ddagger$~\citep{Kwiatowski13OnTheFly},~$\dagger\dagger$~\citep{Liang13Learning}.\label{tab:semantic-results}}
\end{table}
\end{comment}

\defcitealias{Zettlemoyer07Online}{ZC07}
\defcitealias{Kwiatowski13OnTheFly}{KCAZ13}
\defcitealias{Liang13Learning}{LJK13}
\defcitealias{Wang14Morpho}{WKZ14}
\defcitealias{Popescu03Databases}{PEK03}
\defcitealias{Dong16Semantic}{DL16}
\defcitealias{Zhao15Incremental}{ZH15}

\begin{table*}[t]
\centering
\begin{tabular}{|lc|lc|lc|}

  \multicolumn{2}{c}{\Atis}
& \multicolumn{2}{c}{\Geo}
& \multicolumn{2}{c}{\Jobs}
\\
\hline

% & & & & & \\

  System & Accuracy
& System & Accuracy
& System & Accuracy
\\

\hline

  \citetalias{Zhao15Incremental} & 84.2
& \citetalias{Zhao15Incremental} & 88.9
& \citetalias{Zhao15Incremental} & 85.0
\\

  \citetalias{Zettlemoyer07Online} & 84.6
& \citetalias{Kwiatowski13OnTheFly} & 89.0
& \citetalias{Popescu03Databases} & 88.0
\\

  \citetalias{Wang14Morpho} & {\bf 91.3}
& \citetalias{Wang14Morpho} & {\bf 90.4}
& \citetalias{Liang13Learning} & 90.7
\\

\hline

  \citetalias{Dong16Semantic} & 84.6
& \citetalias{Dong16Semantic} & 87.1
& \citetalias{Dong16Semantic} & 90.0
\\

  \ASN & 85.3
& \ASN & 85.7
& \ASN & {\bf 91.4}
\\

  ~~ + \SupAtt & 85.9
& ~~ + \SupAtt & 87.1
& ~~ + \SupAtt & {\bf 92.9}
\\

\hline

\end{tabular}
\caption{Accuracies for the semantic parsing tasks. \ASN denotes our abstract syntax network framework. \SupAtt refers to the supervised attention mentioned in Section~\ref{subsec:attention}. \label{tab:semantic-results}}
\end{table*}

\subsection{Semantic parsing}

\paragraph{Data} We use three semantic parsing datasets: \Jobs, \Geo, and \Atis. All three consist of natural language queries paired with a logical representation of their denotations. \Jobs consists of 640 such pairs, with Prolog-style logical representations, while \Geo and \Atis consist of 880 and 5,410 such pairs, respectively, with $\lambda$-calculus logical forms. We use the same training-test split as~\citet{Zettlemoyer05Semantic} for \Jobs and \Geo, and the standard training-development-test split for \Atis. We use the preprocessed versions of these datasets made available by~\citet{Dong16Semantic}, where text in the input has been lowercased and stemmed using NLTK~\citep{Bird09NLP}, and matching entities appearing in the same input-output pair have been replaced by numbered abstract identifiers of the same type.

\paragraph{Evaluation} We compute accuracies using tree exact match for evaluation. Following the publicly released code of~\citet{Dong16Semantic}, we canonicalize the order of the children within conjunction and disjunction nodes to avoid spurious errors, but otherwise perform no transformations before comparison.

\subsection{Code generation}

\paragraph{Data} We use the \HS dataset introduced by~\citet{Ling16Code}, which consists of 665 cards paired with their implementations in the open-source Hearthbreaker engine.\footnote{Available online at \url{https://github.com/danielyule/hearthbreaker}.}
Our training-development-test split is identical to that of~\citet{Ling16Code}, with split sizes of 533, 66, and 66, respectively.

Cards contain two kinds of components: textual components that contain the card's name and a description of its function, and categorical ones that contain numerical attributes (attack, health, cost, and durability) or enumerated attributes (rarity, type, race, and class). The name of the card is represented as a sequence of characters, while its description consists of a sequence of tokens split on whitespace and punctuation. All categorical components are represented as single-token sequences.

\paragraph{Evaluation}

For direct comparison to the results of~\citet{Ling16Code}, we evaluate our predicted code based on exact match and token-level BLEU relative to the reference implementations from the library. We additionally compute node-based precision, recall, and F1 scores for our predicted trees compared to the reference code ASTs. Formally, these scores are obtained by defining the intersection of the predicted and gold trees as their largest common tree prefix. 

\subsection{Settings} For each experiment, all feedforward and LSTM hidden dimensions are set to the same value. We select the dimension from \{30, 40, 50, 60, 70\} for the smaller \Jobs and \Geo datasets, or from \{50, 75, 100, 125, 150\} for the larger \Atis and \HS datasets. The dimensionality used for the inputs to the encoder is set to 100 in all cases. We apply dropout to the non-recurrent connections of the vertical and horizontal LSTMs, selecting the noise ratio from \{0.2, 0.3, 0.4, 0.5\}. All parameters are randomly initialized using Glorot initialization~\citep{Glorot10Understanding}.

We perform 200 passes over the data for the \Jobs and \Geo experiments, or 400 passes for the \Atis and \HS experiments. Early stopping based on exact match is used for the semantic parsing experiments, where performance is evaluated on the training set for \Jobs and \Geo or on the development set for \Atis. Parameters for the \HS experiments are selected based on development BLEU scores. In order to promote generalization, ties are broken in all cases with a preference toward higher dropout ratios and lower dimensionalities, in that order.

Our system is implemented in Python using the DyNet neural network library~\citep{Neubig17Dynet}. We use the Adam optimizer~\citep{Kingma14Adam} with its default settings for optimization, with a batch size of 20 for the semantic parsing experiments, or a batch size of 10 for the \HS experiments.

\subsection{Results}

\begin{table}[h]
\centering
\begin{tabular}{|lccc|}
\hline
System & Accuracy & BLEU & F1 \\
\hline
\Nearest & 3.0 & 65.0 & 65.7 \\ 
\LPN & 6.1 & 67.1 & -- \\
\ASN & {\bf 18.2} & {\bf 77.6} & {\bf 72.4} \\
~~ + \SupAtt & {\bf 22.7} & {\bf 79.2} & {\bf 75.6} \\
\hline
\end{tabular}
\caption{Results for the \HS task. \SupAtt refers to the system with supervised attention mentioned in Section~\ref{subsec:attention}. \LPN refers to the system of~\citet{Ling16Code}. Our nearest neighbor baseline \Nearest follows that of~\citet{Ling16Code}, though it performs somewhat better; its nonzero exact match number stems from spurious repetition in the data.}
\label{tab:hearthstone-results}
\end{table}

Our results on the semantic parsing datasets are presented in Table~\ref{tab:semantic-results}. Our basic system achieves a new state-of-the-art accuracy of 91.4\% on the \Jobs dataset, and this number improves to 92.9\% when supervised attention is added. On the \Atis and \Geo datasets, we respectively exceed and match the results of \citet{Dong16Semantic}. However, these fall short of the previous best results of 91.3\% and 90.4\%, respectively, obtained by \citet{Wang14Morpho}. This difference may be partially attributable to the use of typing information or rich lexicons in most previous semantic parsing approaches~\citep{Zettlemoyer07Online,Kwiatowski13OnTheFly,Wang14Morpho,Zhao15Incremental}.

%Our basic system achieves new state-of-the-art accuracies on the \Atis and \Jobs datasets of 85.3\% and 91.4\%, respectively, improving over the prior best results of 84.6\%~\citep{Zettlemoyer07Online} and 90.7\%~\citep{Liang13Learning}. When supervised attention is incorporated, these scores further improve to 85.9\% and 92.9\%. We match the results of \citet{Dong16Semantic} on the \Geo dataset, achieving a top accuracy of 87.1\%. However, this falls slightly short of the previous best result of 89.0\% obtained by \citet{Kwiatowski13OnTheFly}.

On the \HS dataset, we improve significantly over the initial results of \citet{Ling16Code} across all evaluation metrics, as shown in Table~\ref{tab:hearthstone-results}. On the more stringent exact match metric, we improve from 6.1\% to 18.2\%, and on token-level BLEU, we improve from 67.1 to 77.6. When supervised attention is added, we obtain an additional increase of several points on each scale, achieving peak results of 22.7\% accuracy and 79.2 BLEU.

\begin{figure}[!h]
\centering
\setlength\tabcolsep{2pt}
\begin{tabular}{cp{0.6\linewidth}}
\raisebox{-\totalheight}{\includegraphics[trim={0cm 0cm 0.5cm 0.6cm},clip,width=0.4\linewidth]{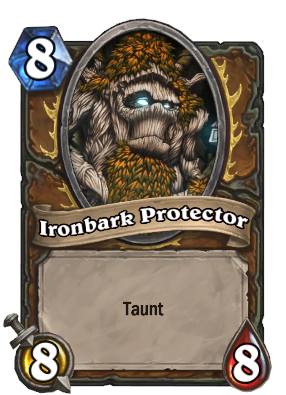}} &
\begin{minted}[fontsize=\scriptsize]{python}
class IronbarkProtector(MinionCard):
  def __init__(self):
    super().__init__(
      'Ironbark Protector', 8,
      CHARACTER_CLASS.DRUID,
      CARD_RARITY.COMMON)
  def create_minion(self, player):
    return Minion(
      8, 8, taunt=True)
\end{minted}
\end{tabular}
\caption{Cards with minimal descriptions exhibit a uniform structure that our system almost always predicts correctly, as in this instance.}
\label{fig:hearthstone-easy}
\end{figure}

\begin{figure}[!h]
\centering
\setlength\tabcolsep{2pt}
\begin{tabular}{cp{0.6\linewidth}}
\raisebox{-\totalheight}{\includegraphics[trim={0cm 0cm 0.5cm 0.6cm},clip,width=0.4\linewidth]{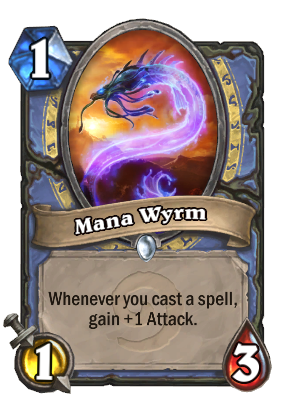}} &
\begin{minted}[fontsize=\scriptsize]{python}
class ManaWyrm(MinionCard):
  def __init__(self):
    super().__init__(
      'Mana Wyrm', 1,
      CHARACTER_CLASS.MAGE,
      CARD_RARITY.COMMON)
  def create_minion(self, player):
    return Minion(
      1, 3, effects=[
        Effect(
          SpellCast(),
          ActionTag(
            Give(ChangeAttack(1)),
                 SelfSelector()))
      ])
\end{minted}
\vspace{-50pt}
\end{tabular}
\caption{For many cards with moderately complex descriptions, the implementation follows a functional style that seems to suit our modeling strategy, usually leading to correct predictions.}
\label{fig:hearthstone-medium}
\end{figure}

\begin{figure*}[t]
\centering
\setlength\tabcolsep{2pt}
\begin{tabular}{cp{0.375\linewidth}p{0.375\linewidth}}
\raisebox{-\totalheight}{\includegraphics[trim={0cm 0cm 0.5cm 0.6cm},clip,width=0.25\linewidth]{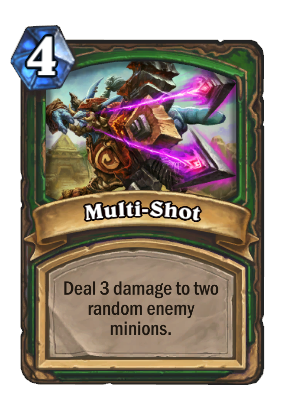}}
&
\begin{minted}[fontsize=\scriptsize]{python}
class MultiShot(SpellCard):
  def __init__(self):
    super().__init__(
      'Multi-Shot', 4,
      CHARACTER_CLASS.HUNTER,
      CARD_RARITY.FREE)
  def use(self, player, game):
    super().use(player, game)
    targets = copy.copy(
                game.other_player.minions)
    for i in range(0, 2):
      target = game.random_choice(targets)
      targets.remove(target)
      target.damage(
        player.effective_spell_damage(3), 
        self)
  def can_use(self, player, game):
    return (
      super().can_use(player, game) and
      (len(game.other_player.minions) >= 2))
\end{minted}
\vspace{-30pt}
&
\begin{minted}[fontsize=\scriptsize]{python}
class MultiShot(SpellCard):
  def __init__(self):
    super().__init__(
      'Multi-Shot', 4, 
      CHARACTER_CLASS.HUNTER, 
      CARD_RARITY.FREE)
  def use(self, player, game):
    super().use(player, game)
    minions = copy.copy(
                game.other_player.minions)
    for i in range(0, 3):
      minion = game.random_choice(minions)
      minions.remove(minion)
  def can_use(self, player, game):
    return (
      super().can_use(player, game) and
      len(game.other_player.minions) >= 3)
\end{minted}
\end{tabular}
\caption{Cards with nontrivial logic expressed in an imperative style are the most challenging for our system. In this example, our prediction comes close to the gold code, but misses an important statement in addition to making a few other minor errors. \emph{(Left)} gold code; \emph{(right)} predicted code.}
\label{fig:hearthstone-hard}
\end{figure*}

\subsection{Error Analysis and Discussion}

As the examples in Figures~\ref{fig:hearthstone-easy}-\ref{fig:hearthstone-hard} show, classes in the \HS dataset share a great deal of common structure. As a result, in the simplest cases, such as in Figure~\ref{fig:hearthstone-easy}, generating the code is simply a matter of matching the overall structure and plugging in the correct values in the initializer and a few other places. In such cases, our system generally predicts the correct code, with the exception of instances in which strings are incorrectly transduced. Introducing a dedicated copying mechanism like the one used by \citet{Ling16Code} or more specialized machinery for string transduction may alleviate this latter problem.

The next simplest category of card-code pairs consists of those in which the card's logic is mostly implemented via nested function calls. Figure~\ref{fig:hearthstone-medium} illustrates a typical case, in which the card's effect is triggered by a game event (a spell being cast) and both the trigger and the effect are described by arguments to an \texttt{Effect} constructor. Our system usually also performs well on instances like these, apart from idiosyncratic errors that can take the form of under- or overgeneration or simply substitution of incorrect predicates.

Cards whose code includes complex logic expressed in an imperative style, as in Figure~\ref{fig:hearthstone-hard}, pose the greatest challenge for our system. Factors like variable naming, nontrivial control flow, and interleaving of code predictable from the description with code required due to the conventions of the library combine to make the code for these cards difficult to generate. In some instances (as in the figure), our system is nonetheless able to synthesize a close approximation. However, in the most complex cases, the predictions deviate significantly from the correct implementation. 

In addition to the specific errors our system makes, some larger issues remain unresolved. Existing evaluation metrics only approximate the actual metric of interest: functional equivalence. Modifications of BLEU, tree F1, and exact match that canonicalize the code---for example, by anonymizing all variables---may prove more meaningful. Direct evaluation of functional equivalence is of course impossible in general~\citep{Sipser2006Introduction}, and practically challenging even for the \HS dataset because it requires integrating with the game engine. 

Existing work also does not attempt to enforce {\it semantic} coherence in the output. Long-distance semantic dependencies, between occurrences of a single variable for example, in particular are not modeled. Nor is well-typedness or executability. Overcoming these evaluation and modeling issues remains an important open problem.

%\paragraph{Over- and Undergeneration}
%
%Although our model is constrained to produce the correct children for composite nodes, variable-length child sequences are generated one subtree at a time by a horizontal LSTM. As a result, our decoder exhibits the same over- and undergeneration pathologies observed in work on neural sequence prediction~\citep{Tu16Modeling}. One potential solution is attention feeding~\citep{Luong15Effective}, where attention weights are threaded through the decoding process. An alternate strategy that might also be effective would be maintaining a soft checklist of items that should be incorporated into the output~\citep{Kiddon16Globally}.

%\paragraph{Identifier Errors}
%
%Novel strings not encountered at training time such as the name of a class must be generated character by character. Although the transformations required primarily consist of removing spaces and punctuation and performing camelcasing, our system still makes mistakes during string generation. Introducing a dedicated copying mechanism like the one used by \citet{Ling16Code} or more specialized machinery for string transduction may alleviate this problem.

\section{Conclusion}

%We introduce abstract syntax networks (ASN), a framework for mapping unstructured or partially structured inputs to well-formed, executable outputs. ASNs employ a modular decoder, each of whose modules corresponds to a construct in the output space grammar and is invoked when that construct is used by the decoder. We apply our approach to the \HS code generation dataset and three benchmark semantic parsing datasets, in most cases achieving state-of-the-art performance.

ASNs provide a modular encoder-decoder architecture that can readily accommodate a variety of tasks with structured output spaces. They are particularly applicable in the presence of recursive decompositions, where they can provide a simple decoding process that closely parallels the inherent structure of the outputs. Our results demonstrate their promise for tree prediction tasks, and we believe their application to more general output structures is an interesting avenue for future work.

\section*{Acknowledgments}

MR is supported by an NSF Graduate Research Fellowship and a Fannie and John Hertz Foundation Google Fellowship. MS is supported by an NSF Graduate Research Fellowship.

%\bibliography{acl2017}
\bibliography{acl2017}
\bibliographystyle{acl_natbib}

\appendix

\newpage %~\newpage

\section{Appendix}

\begin{figure}[h]
\begin{minipage}[t]{0.5\linewidth}
\footnotesize
\begin{verbatim}
expr
  = Apply(pred predicate, arg* arguments)
  | Not(expr argument)
  | Or(expr left, expr right)
  | And(expr* arguments)

arg
  = Literal(lit literal)
  | Variable(var variable)
\end{verbatim}
\end{minipage}
\caption{The Prolog-style grammar we use for the \Jobs task. \label{fig:prolog-grammar}}
\end{figure}

\begin{figure}[h]
\begin{minipage}[t]{0.5\linewidth}
\footnotesize % smaller: \scriptsize, \tiny
\begin{verbatim}
expr 
  = Variable(var variable)
  | Entity(ent entity)
  | Number(num number)
  | Apply(pred predicate, expr* arguments)
  | Argmax(var variable, expr domain, expr body)
  | Argmin(var variable, expr domain, expr body)
  | Count(var variable, expr body)
  | Exists(var variable, expr body)
  | Lambda(var variable, var_type type, expr body)
  | Max(var variable, expr body)
  | Min(var variable, expr body)
  | Sum(var variable, expr domain, expr body) 
  | The(var variable, expr body) 
  | Not(expr argument) 
  | And(expr* arguments) 
  | Or(expr* arguments) 
  | Compare(cmp_op op, expr left, expr right) 

cmp_op = Equal | LessThan | GreaterThan
\end{verbatim}
\end{minipage}
\caption{The $\lambda$-calculus grammar used by our system. \label{fig:lambda-grammar}}
\end{figure}

\end{document}